\documentclass[preprint,12pt]{elsarticle}

\usepackage{cite}
\usepackage{graphicx}
\usepackage{geometry}
\usepackage{booktabs}
\usepackage{amsfonts}
\usepackage{amsmath}
\usepackage{float}
\usepackage{subcaption}

\journal{Neural Networks}

\graphicspath{ {images/} }

\begin{document}

\begin{frontmatter}

\title{Improved robustness of reinforcement learning policies upon conversion to spiking neuronal network platforms applied to Atari Breakout game}

\author[a]{Devdhar Patel\corref{cor1}}
\ead{devdharpatel@cs.umass.edu}

\author[a]{Hananel Hazan}

\author[a]{Daniel J. Saunders}

\author[a]{Hava T. Siegelmann}

\author[a,b]{Robert Kozma\corref{cor1}} 
\ead{rkozma@cs.umass.edu}


\address[a]{Biologically Inspired Neural and Dynamical Systems Laboratory (BINDS)\\
College of Computer and Information Sciences, 140 Governors Drive\\
University of Massachusetts Amherst, Amherst, MA 01003, USA\\
.}

\address[b]{Center for Large-Scale Integrated Optimization and Networks (CLION)\\
Department of Mathematical Sciences, 373 Dunn Hall\\
University of Memphis, Memphis, TN 38152, USA}

\begin{abstract}
Deep Reinforcement Learning (RL) demonstrates excellent performance on tasks that can be solved by trained policy. 
It plays a dominant role among cutting-edge machine learning approaches using multi-layer Neural networks (NNs). 
At the same time, Deep RL suffers from high sensitivity to noisy, incomplete, and misleading input data. 
Following biological intuition, we involve Spiking Neural Networks (SNNs) to address some deficiencies of deep RL solutions. 
Previous studies in image classification domain demonstrated that standard NNs (with ReLU nonlinearity) trained using supervised learning can be converted to SNNs with negligible deterioration in performance. In this paper, we extend those conversion results to the domain of Q-Learning NNs trained using RL. We provide a proof of principle of the conversion of standard NN to SNN. In addition, we show that the SNN has improved robustness to occlusion in the input image. Finally, we introduce results with converting full-scale Deep Q-network to SNN, paving the way for future research to robust Deep RL applications.

\end{abstract}

\cortext[cor1]{Corresponding author}
\end{frontmatter}

\section{Introduction}

Among the giants of neural networks and brain science, Stephen Grossberg has a truly unique and overarching legacy in the research fields encompassing the physiology and mathematical modeling of neural processing and brain functions. From the enormous amount of his influential research achievements produced for over 60 years, here we emphasize his results on laminar cortical models of spiking neurons and laminar computing. 
The Synchronous Matching Adaptive Resonance Theory (SMART) model is a groundbreaking study employing synchronous oscillations in spiking neurons to describe attentive learning in thalamocortical circuits \citep{GrossbergVersace08}. The laminar computing principle later has been extended to modeling visual cortical processing and the formation of visual percepts \citep{Leveille10, CaoGrossberg12}. Grossberg's laminar computing approach has important impact on not only computational models, but also on hardware developments, and it provides a blueprint for advanced chip designs to implement various machine learning tasks, such as IBM's TrueNorth \citep{Pedroni16}, Intel's Loihi \citep{Davies18}, and SpiNNaker at Manchester, UK \citep{Furber14}. These chips employ spiking neural networks, which provide the energy efficiency required for the future dynamical development of {\em sustainable AI} and brain-inspired technologies.

Recent advancements in AI and Machine Learning (ML) have astonishing results surpassing human performance in various testbeds, including ATARI games \citep{mnih2015humanlevel, Hasselt:2016:DRL:3016100.3016191, pmlr-v48-wangf16}. These successes have lead to enormous interest in AI and neural networks, including Deep Reinforcement Learning (RL). However, rigorous analysis showed that deep RL is susceptible to random and malicious perturbations in the inputs, related to adversarial AI \citep{Huang2017AdversarialAO}. 
A consequence of the applied gradient descent algorithm is that the trained agent learns to focus on a few sensitive areas. However, the performance of the RL agent deteriorates when these areas are occluded or perturbed. Moreover, there is evidence that the policies learned by the networks in deep RL algorithms do not generalize well; the performance of the agent deteriorates when it encounters a state that it has not seen before, even if it is similar to other experienced states \citep{witty2018measuring}.

Biological systems tend to be very noisy by nature \citep{noise_bio,noise_bio2}, still they operate well even under harsh conditions that affect their input and internal state. Spiking Neural Networks (SNNs) are considered to be closer to biological neurons due to their event-based nature; they are often termed the third generation of neural networks \citep{Maass1996NetworksOS}. A spike is the quantification of the internal and external processes involving the neuron. 
The individual neurons operating with spikes serve as microscopic bottlenecks, which have the ability to sustain low intermittent noise and do not transmit sub-threshold noise to their neighbors. Moreover, populations of spiking neurons in a network can mitigate the impact of noise even further due to their collective effect and their architectural connectivity \citep{HAZAN20121597}. Following biological intuition, we involve SNNs to enhance the benefits of deep RL solutions.

An important potential advantage of SNNs is their energy efficiency. 
Due to the binary, event-based nature, SNNs can support energy utilization that is more efficient than the one provided by traditional neural networks, especially when implemented on neuromorphic hardware \citep{Mart2016EnergyEfficientNC}. Among the various hardware solutions, memristor technology demonstrated great promise for neuromorphic computing \citep{Kozma12,Srinivasa12,Bichler13}.
In recent years, we witnessed the proliferation of 
neuromorphic hardware platforms, such as IBM TrueNorth and Intel Loihi \citep{Benjamin14,Pedroni16,Davies18}.


It is difficult to train spiking neurons using backpropagation due to the non-differentiable nature of the spike dynamics \citep{10.3389/fnins.2018.00774}.
One important direction of recent work with SNNs has focused on adjusting backpropagation to the event-based nature of spiking neuron activation \citep{NIPS2018_7417, 10.3389/fnins.2018.00331,Zenke18}. An alternative approach aimed at biologically inspired local learning rules, e.g., spike-timing-dependent plasticity (STDP) to train the network \citep{bengio2015stdp, 10.3389/fncom.2015.00099,non_linear_snn,10.3389/fncom.2018.00024,Hazan18,Davies18}. 

In this work, we explore a different approach, in which no learning takes place in the SNN at all, rather the weights obtained by training a ReLU NN are converted to the SNN having the same structure. This idea may sound either trivial or crazy, still there is ample of evidence that it indeed works. The idea of converting Convolutional NNs (CNNs) to SNNs with the aim of processing inputs from event-based sensors was first introduced in Perez-Carrasco et al.\citep{6497055} Cao et al.\citep{Cao2015} observed that the the activations of ReLU neurons can be mapped to the frequency of spikes produced by the spiking neurons and reported good performance on computer vision benchmarks. 

Diehl et al.\citep{7280696} proposed a method of weight normalization that re-scales the weights of the SNN to reduce the errors due to excessive spiking or due to sparse spiking of the neurons. They also demonstrated a near lossless conversion of ReLU NNs to SNNs for the MNIST classification task. Rueckauer and colleagues identified spiking equivalents of a variety of common operations used in deep convolutional networks like max-pooling, softmax, batch-normalization, and inception modules \citep{rueckauer2016theory, 10.3389/fnins.2017.00682}). This allowed them to convert popular CNN architectures like VGG-16 to SNN \citep{vgg-16}, Inception-V3\citep{Inception-V3}, BinaryNet\citep{BinaryNet}. They achieved near lossless conversion in these networks. All these efforts aimed at classification tasks, while according to our knowledge, there has not been previous work on the conversion of Deep Q-networks to SNNs. 

The combination of RL and spiking neurons is a natural choice, since animals learn to perform certain tasks using semi-supervised and reinforcement learning. Moreover, there is evidence that biological neurons learn using evaluative feedback from neurotransmitters such as dopamine\citep{Wang2018PrefrontalCA}, e.g., in the postulated dopamine reward prediction-error signal \citep{Schultz2016DopamineRP}. However, since spiking neurons are fundamentally different from ReLU artificial neurons, it is not clear if SNNs can address machine learning in RL domain. This raises the questions: Do SNNs have the capability to represent the same functions as ReLU NNs? To be more specific, can SNNs represent complex policies that can successfully play Atari games? 

We answer these questions by demonstrating that ReLU NNs trained using RL algorithms can be converted to SNN without deterioration of the performance on the RL task when playing Atari Breakout game. Furthermore, we show that the converted SNN is more robust to input perturbations than the original NN. Finally, we demonstrate that full-sized Deep Q-Network (DQN) \citep{mnih2015humanlevel} can be converted to SNN and maintain its better than human performance, paving the way for future research in robustness and RL with SNNs.
Results presented in this paper has been produced using the open source {\texttt BindsNET} spiking neural networks library, available on Github \texttt{https://github.com/Hananel-Hazan/bindsnet}.

\section{Background}
\subsection{Arcade learning environment}
The Arcade learning environment (ALE) \citep{bellemare13arcade} is a platform that enables researchers to test their algorithms on over 50 Atari 2600 games. The agent sees the environment through image frames of the game, interacts with the environment with 18 possible actions, and receives feedback in the form of the change in the game score. The games were designed for humans and thus are free from experimenter bias. The games span many different genres that require the agent/algorithm to generalize over various tasks, difficulty levels, and timescales. ALE thus has become a popular test-bed for reinforcement learning \citep{mnih2015humanlevel}.

\begin{figure}[ht]
\vskip 0in
\begin{center}
\centerline{\includegraphics[width=4cm, height=6.4cm]
{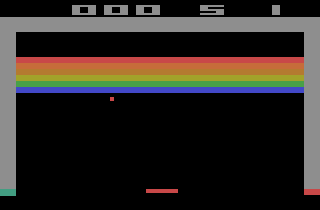}}
\caption{Screenshot of Atari 2600 Breakout game. The ball bounces between the wall (lines of colored bricks) and the paddle (red bar at the bottom).}
\label{fig:breakout}
\end{center}
\vskip -0.2in
\end{figure}
\textit{Breakout:} We demonstrate our results on the game of Breakout; Figure \ref{fig:breakout} shows a frame of the game. Breakout is similar to the popular game Pong. The player controls a paddle at the bottom of the screen; see red bar at the bottom of Figure \ref{fig:breakout}.  There are rows of colored bricks on the upper part of the screen. A ball bounces in between the bricks and the player controlled paddle. There are 4 possible actions: move left, move right, do not move, and fire. If the ball hits a brick, the brick breaks and the score of the game is increased. However, if the ball falls below the paddle, the player loses a life. The game starts with five lives, and the player/agent is supposed to break all the bricks before they run out of lives. 

\subsection{Deep Q-Networks}
Reinforcement learning algorithms train a policy $\pi$ to maximize the expected cumulative reward received over time. Formally, this process is modeled as a Markov decision process (MDP). Given a state-space $\mathcal{S}$ and an action-space $\mathcal{A}$, the agent starts in an initial state $s_0 \in \mathcal{S_0}$ from a set of possible start states $S_0 \in \mathcal{S}$. At each time-step $t$, starting from $t=0$, the agent takes an action $a_t$ to transition from $s_t$ to $s_{t+1}$. The probability of transitioning from state $s$ to state $s'$ by taking action $a$ is given by the transition function $P(s,a,s')$. The reward function $R(s,a)$ defines the expected reward received by the agent after taking action $a$ on state $s$. 

A policy $\pi$ is defined as the conditional distribution of actions given the state $\pi(s,a) = Pr(A_t = a | S_t = s)$. The Q-value or action-value of a state-action pair for a given policy, $q^\pi(s,a)$, is the expected return following policy $\pi$ after taking the action $a$ from state $s$. 
\begin{equation}
    q^\pi(s,a) = \mathbb{E}[\sum_{k=0}^\infty \gamma^k R_{t+k}|S_t=s, A_t = a, \pi]
\end{equation}
where $\gamma$ is the discount factor. The action-value function follows a Bellman equation that can be written as:
\begin{equation}
    q^\pi(s_t,a_t) = r_t + \gamma \max_{a_{t+1}}q^\pi(s_{t+1}, a_{t+1})
\end{equation}

Many widely used reinforcement learning algorithms first approximate the Q-value and then select the policy that maximizes the Q-value at each step to maximize returns \citep{suttonbarto}. Deep Q-networks (DQN) \citep{mnih2015humanlevel} are one such algorithm that uses deep artificial neural networks to approximate the Q-value. The neural network can learn policies from the pixels of the screen and the game score. It has been shown that DQN surpasses human performance on many of the Atari games.

\begin{figure}[ht]
\vskip 0in
\begin{center}
\centerline{\includegraphics[width=\columnwidth]{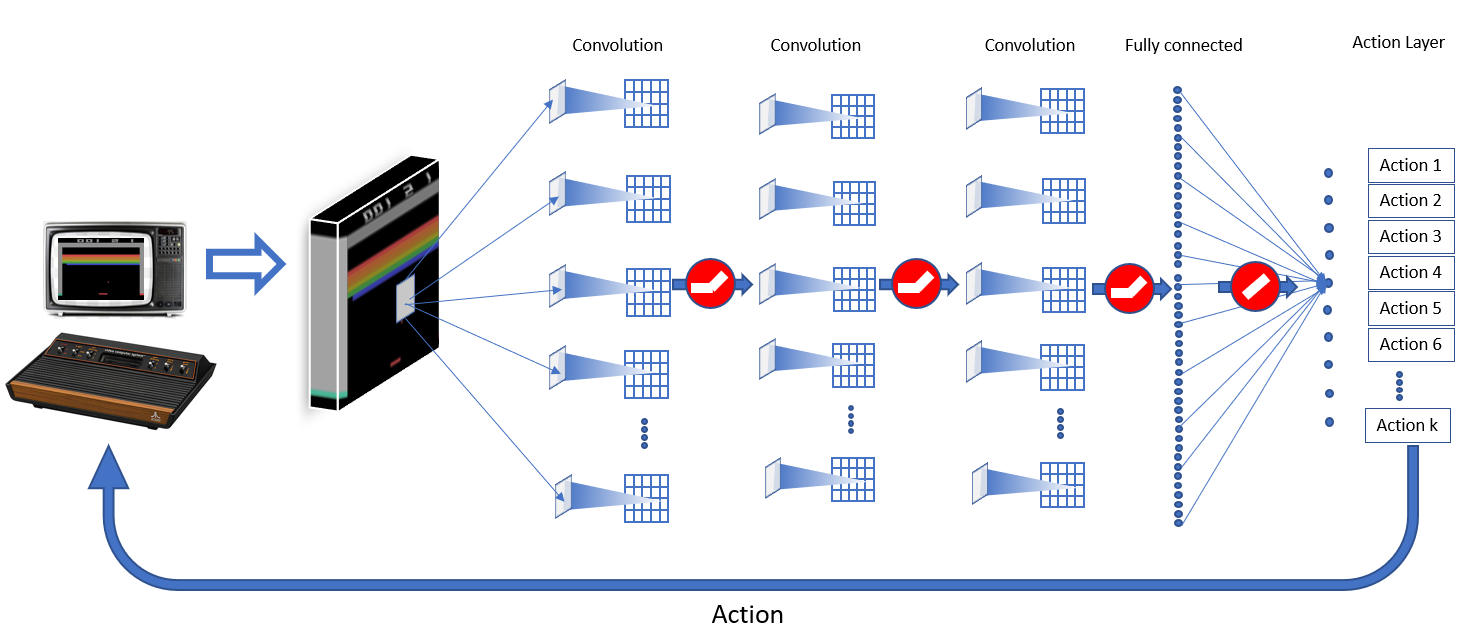}}
\caption{Architecture of Deep Q-networks; following  Mnih et al.\citep{mnih2015humanlevel} ; ReLU nonlinear units are emphasized by red circles. }
\label{fig:Full_Size_CNN}
\end{center}
\vskip -0.2in
\end{figure}

\subsection{Spiking neurons}
SNNs may use any of the various neuron models \citep{spiking, spiking2}. Here, we introduce four different types of spiking neurons. We use the following notations to describe the dynamics of the neurons:
$v(t)$ is the time-dependent membrane potential voltage;
$v_{rest}$ is the resting membrane potential;
$v_{thresh}$ is the firing threshold of the neuron;
$\tau$ is the time constant of the neuron dynamics.

\begin{enumerate}
    \item \textit{Integrate-and-fire (IF) neuron:} The IF neuron is the simplest form of spiking neuron models. The neuron simply integrates input until the membrane potential $v(t)$ exceeds the voltage threshold $v_{threshold}$ and a spike is generated. Once the spike is generated, the membrane potential is reset to $v_{reset}$.
	\begin{equation}
	\tau \frac{dv(t)}{dt} = \sum\limits_{i=1}^n W_{i} * Input_{i} .
	\end{equation}
    
    \item \textit{Subtractive Integrate-and-fire (SubIF) neuron:} The SubLIF neuron behaves similar to the IF neuron with one small change, when the membrane potential voltage exceeds threshold value, the neuron emits a spike and resets its membrane voltage to  $v_{reset} + (v(t) - v_{threshold})$ \citep{SubLIF,SubLIF2,10.3389/fnins.2017.00682}. By adding the overshoot voltage the neuron "remembers" the excessive voltage from the last spike and will be more prone to be excited with the next incoming inputs. This reduces the information lost when spiking in SNN is converted from ReLU NN.

    \item \textit{Leaky integrate-and-fire (LIF) neuron:} The LIF neuron behaves similarly to the IF neuron. However, for every time-step that its membrane potential is above the resting potential, the neuron leaks a constant amount of current:
	\begin{equation}
		\tau \frac{dv(t)}{dt} = - (v_{t} - v_{rest}) + \sum\limits_{i=1}^n W_{i} * Input_{i} .
	\end{equation}
    
    \item \textit{Stochastic leaky integrate-and-fire neuron:} The stochastic LIF neuron is based on the LIF neuron. However, the neuron may spike if its membrane potential is below the threshold with probability proportional to its membrane potential (escape noise). The escape noise ($\sigma$) is described here:

    \begin{equation}
    \begin{gathered}
    \sigma =
        \begin{cases}
            \textrm{$1/\tau_\sigma\exp(\beta_\sigma(v_{t} - v_{threshold}))$} & \textrm{if less than 1}\\
            1 & \textrm{otherwise}, 
       \end{cases}
    \end{gathered}
    \end{equation}
\end{enumerate}
where $\tau_\sigma$ and $\beta_\sigma$ are positive constant parameters. 
For the spiking models listed above, the neuron enters a refractory period after a spike, during which they are unable to spike or integrate input. In this paper, for simplicity, we ignore the refractory period in the conversion from artificial neurons, and we set both $\tau_\sigma$ and $\beta_\sigma$ to 1. For a complete list of the parameters used for the neurons, see \textit{Supplementary Materials}, Table 2. 

Note that unlike traditional artificial NNs, SNNs need to be simulated for a period of time to produce spike trains and interpret the resulting activity. The simulation is done in discrete time steps. To avoid confusion between the time step of the RL environment and the time step of the SNN, we denote the time step of the RL environment by $t$ and the time step of the SNN by $nt$.

\section{Methods}
\subsection{Inputs}
\subsubsection{Binary input}
First we consider binary pixel inputs, following Mnih et al.\citep{mnih2015humanlevel}. Each state consists of an 80x80 image of binary pixels. The frames from the AI Gym environment are pre-processed to create the state for further analysis.
Each frame from the AI Gym environment is cropped to remove the text above the screen displaying the score and the number of lives left. The image is then re-sized to an 80x80 image and converted to a binary image. The previous frame is then subtracted from the current frame while clamping all the negative values to 0. We then add the most recent four such difference frames to create a state for the RL environment. Thus, a state is an 80x80 binary image containing the movement information of the last four states. 

\subsubsection{Grayscale input}
The binary input described above does not contain information about the direction of the ball movement, which we believe can confuse the agent. To alleviate this problem, we weighted each frame according to time and added them to create the state. The most recent frame has the highest weight, and the least recent frame has the least weights. At time $t$ the state is made up of the sum of the most recent 4 frames as follows:
\begin{equation}
    S_t = F_t * 1 + F_{t-1} * 0.75 + F_{t-2} * 0.5 + F_{t-3} * 0.25 .
\end{equation}

Here $S_t$ and $F_t$ are the state and the frame at time $t$, respectively. 

\subsection{Network architecture}
The networks used in the DQN algorithm for Atari games consist of multiple convolutional layers followed by fully connected layers \citep{mnih2015humanlevel}. 
In this work, we started by testing our methods on a shallow ReLU NN with one hidden layer, in order to demonstrate the feasibility of using weight transfer in RL problems. Then we moved on to full-sized Deep Q-network with the same architecture as Mnih et al. \citep{mnih2015humanlevel}.

\begin{figure}[ht]
\vskip 0in
\begin{center}
\centerline{\includegraphics[width=\columnwidth]{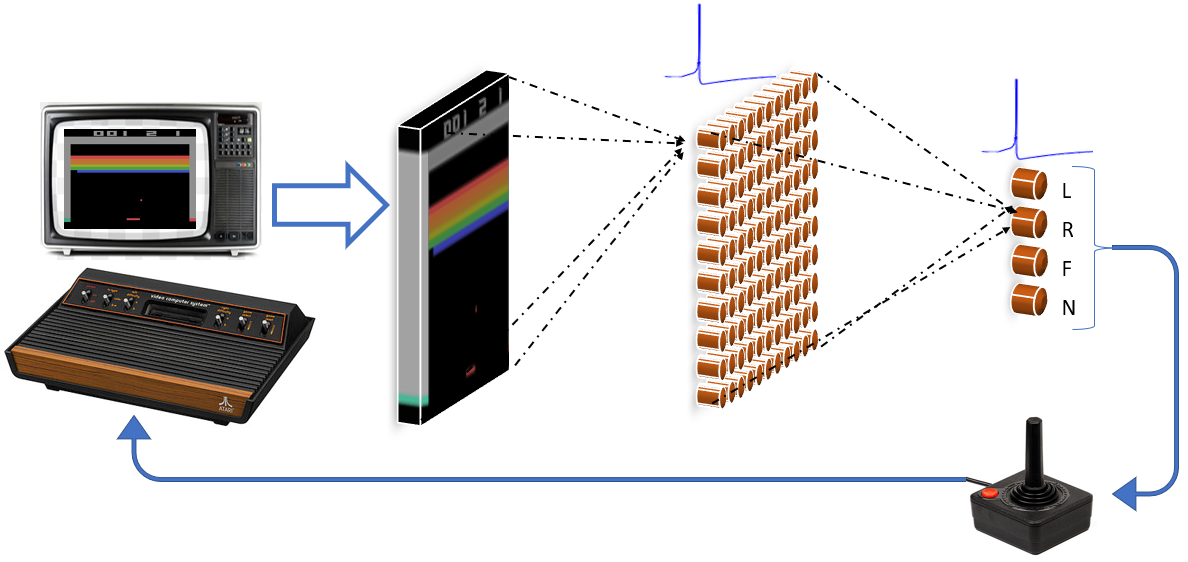}}
\caption{Network architecture: The input to the network consists of an 80x80 image produced by preprocessing the frames of the game. The hidden layer consists of 1000 neurons, the output layer has 4 nodes corresponding to the number of possible actions.}
\label{fig:networks}
\end{center}
\vskip -0.2in
\end{figure}

Figure \ref{fig:networks} shows the network architecture of the shallow SNN. The network architecture of the SNN is the same as the ReLU NN, except that the ReLU nonlinearities of the neurons are replaced by spiking neurons.
The network consists of 80x80 input layer, followed by a fully connected hidden layer with 1000 neurons. The output layer is a fully connected layer with 4 neurons that give the estimate of the optimal action-value of each of the 4 possible actions in the Breakout game.

\subsection{Training by reinforcement Q-learning approach}
We trained the networks using the DQN algorithm \citep{mnih2015humanlevel}. We trained the smaller networks using a replay memory size of 200,000 and initial replay memory size of 50,000.  
We trained the network over 30000 episodes. Each episode refers to one game of breakout with five lives. The episode ends when the agent/player runs out of lives.
The rest of the hyper-parameters we used are same as in Mnih et al.'s\citep{mnih2015humanlevel} work. For a complete list of the hyper-parameters, see \textit{Supplementary Materials}, Table 1. 

\subsection{Conversion of trained ReLU NN to SNN}
The ReLU NN, which has been trained using the DQN algorithm, is converted to SNN. For the converted SNN, the firing frequency of the spiking neurons in the output layer is proportional to the Q-value of the corresponding action.
ReLU NN can be converted to SNN by replacing the ReLU neurons with spiking neurons. 
However, the result of this straight forward conversion may produce a very sparse spiking activity in the network. This is due to the fact that the spiking neurons have a constant positive threshold while ReLU neurons activate at any value above zero.
To address this sparsity issue, the SNN is simulated for a large number of time steps($nt$) for a given input to generate sufficient level of spiking activity, allowing robust estimation of the Q values. 
In our experiments, we simulate the SNN for $nt$ = 500 time-steps, and we repeat this simulation for each input pattern.
In order to expedite the process, we also increase the spiking activity by scaling up the weights. Generally, the weights of deeper layers need to be scaled more than weights of the layers close to the inputs because the network activity becomes sparser in deeper layers. 

Due to the fundamental difference between a spiking neuron and a ReLU neuron, the frequency of the spiking neurons cannot accurately represent the output of equivalent ReLU neurons. This is due to the fact that the spiking neurons can only output discrete spikes, while the ReLU neurons have continuous outputs. The conversion of continuous activity (ReLU) to discrete/spiking activity lies in the very heart of this method. 
The output frequency of the spiking neuron is limited by the choice of membrane potential threshold and simulation time-steps ($nt$). However, we can reduce the error of spiking neurons by scaling the weights of each layer of network and improve the performance of the SNN. We treat the scaling of the weights at each layer as independent parameters to be optimized to achieve high the performance of the network in the Atari game testbed. All the weights of the same layer are scaled by the same factor thus preserving the learned filters. 
The optimal weight scaling parameters can be searched by various methods. Rueckauer et al.\citep{10.3389/fnins.2017.00682} showed a useful approach of scaling by normalizing the weights. Their approach was based on scaling the weights in a way that the output error of the majority of the spiking neurons is minimized. 

There are various alternatives to Rueckauer's method \citep{10.3389/fnins.2017.00682} 
to optimize the transfer from ReLU NN to spiking NNs. For example, \citep{Kaushik19} provides impressive results  in the context of adversarial AI.
In our present study, we explored several ways to search for optimal scaling parameters, including particle swarm optimization (PSO) \citep{clerc:hal-00764996}, and simple exhaustive grid search. Among the studied optimization methods, PSO has produced the best performance, and we briefly summarize it here.
PSO uses particles to evaluate the fitness of various positions inside the search space. The particles inform each other on their previous best positions. Each particle has a velocity attached to it, and the velocity and the position of the particles are updated at each iteration using a set of rules, which allow efficient exploration of the search space. In our approach, the $n^{th}$ dimension of the particle position determines the value, by which the $n^{th}$ layer of the SNN is scaled. The fitness of a particular position is determined by evaluating how well a specific network performs on the game, based on the scaling the coordinates in that position.

PSO-based optimization of the network demonstrates much improved performance on the Atari game w.r.t. Rueckauer et al.'s \citep{10.3389/fnins.2017.00682} method, which reduces the error between the output of spiking neurons and ReLU neurons. In short, PSO acts as a training algorithm for the SNN. PSO is better suited for the networks trained using the DQN algorithm because unlike image classification tasks, which use the cross-entropy loss, the output values of Q-networks do not differ by large values thus making it harder to differentiate when they are discretized in the SNN.

\section{Results}

\subsection{Performance in breakout games using shallow NNs}
Testing SNN based agents in the ALE is a computationally demanding task. 
We simulate spiking neurons using the PyTorch based open source library {\texttt BindsNET}\citep{Hazan18}. {\texttt BindsNET} has the advantage compared to some alternative spiking NN packages of allowing users to leverage GPUs to simulate the SNN and speed up testing. 

\begin{table}[ht]
    \vskip 0.15in
    \begin{center}
    \begin{small}
    \begin{tabular}{lcccr}
        \toprule
        Input & ReLU  & SNN & Stochastic\\
          & NN & with LIF & SNN w/ LIF\\
        \midrule
        \multicolumn{4}{c}{~~~~~~~~~~~~~~~~~~0.05 Epsilon Greedy}\\
        \midrule
        Binary & $5.77 \pm 3.07$ & $6.21 \pm 1.74$ & $7.12 \pm 2.47$\\
        Grayscale & $6.55 \pm 1.53$ & $7.28 \pm 1.79$ &  $7.5 \pm 2.16$\\
        \midrule
        \multicolumn{4}{c}{~~~~~~~~~~~~~~~~~~~Greedy}\\
        \midrule
        Binary & $6.0 \pm 0$ & $5.25 \pm 2.13$ & $7.58 \pm 1.87$\\
        Grayscale & $9.32 \pm 0.63$ & $10.05 \pm 0.68$ &  $8.0 \pm 2.37$\\
        \bottomrule
    \end{tabular}
    \end{small}
    \end{center}
    \caption{Best performance achieved for different inputs and networks. Each value represents an average of 100 episodes.}
    \vskip -0.1in
    \label{table:final}
\end{table}

\begin{figure*}[h]
\vskip 0.2in
\begin{center}
\begin{subfigure}[b]{0.4\linewidth}
\centerline{\includegraphics[width=\linewidth]{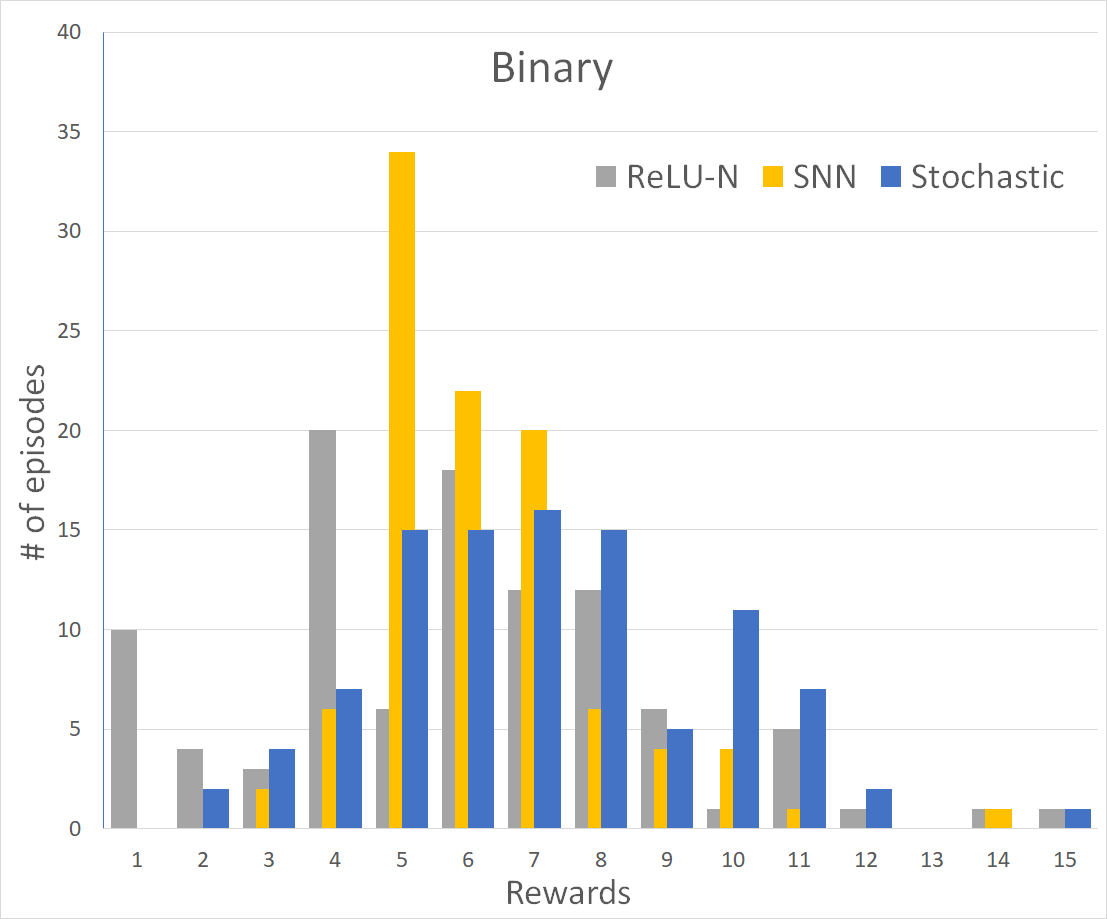}}
\caption{0.05 epsilon greedy binary input}
\end{subfigure}
\begin{subfigure}[b]{0.4\linewidth}
\centerline{\includegraphics[width=\linewidth]{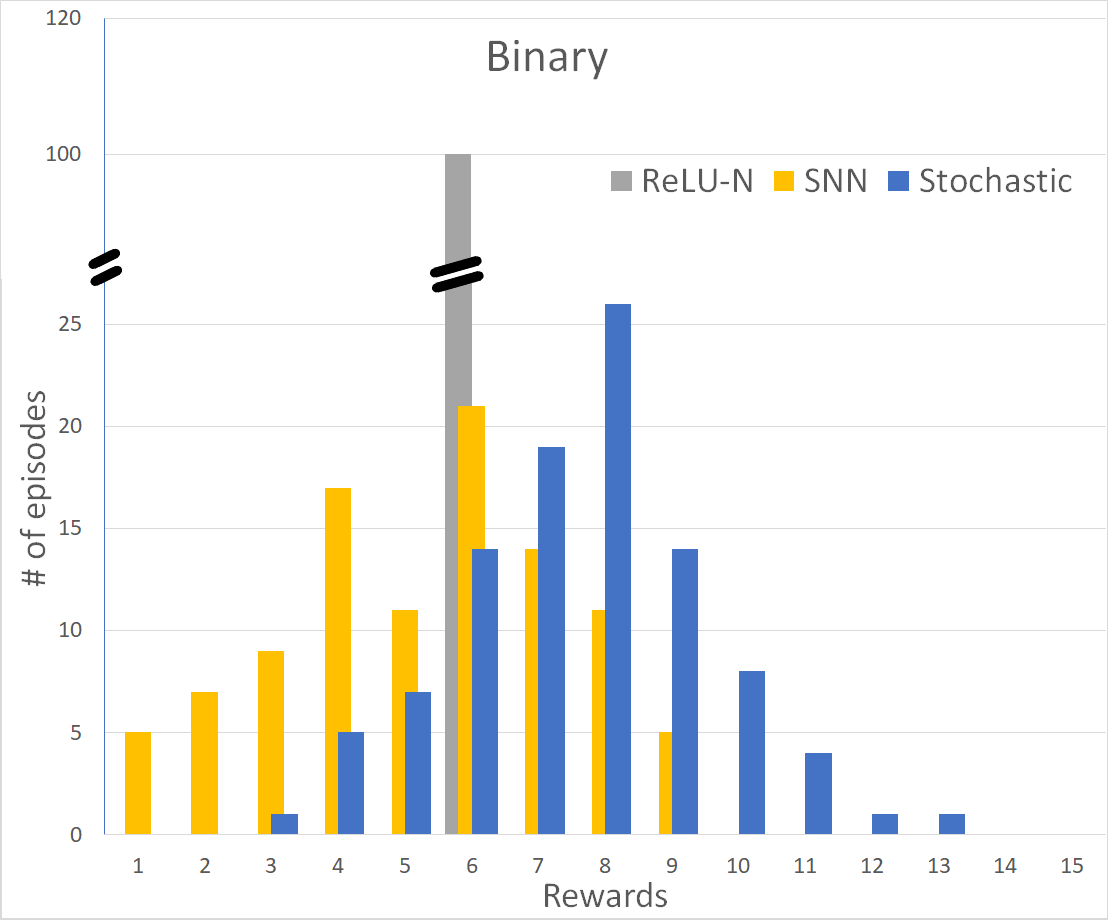}}
\caption{Greedy binary input}
\end{subfigure}
\begin{subfigure}[b]{0.4\linewidth}
\centerline{\includegraphics[width=\linewidth]{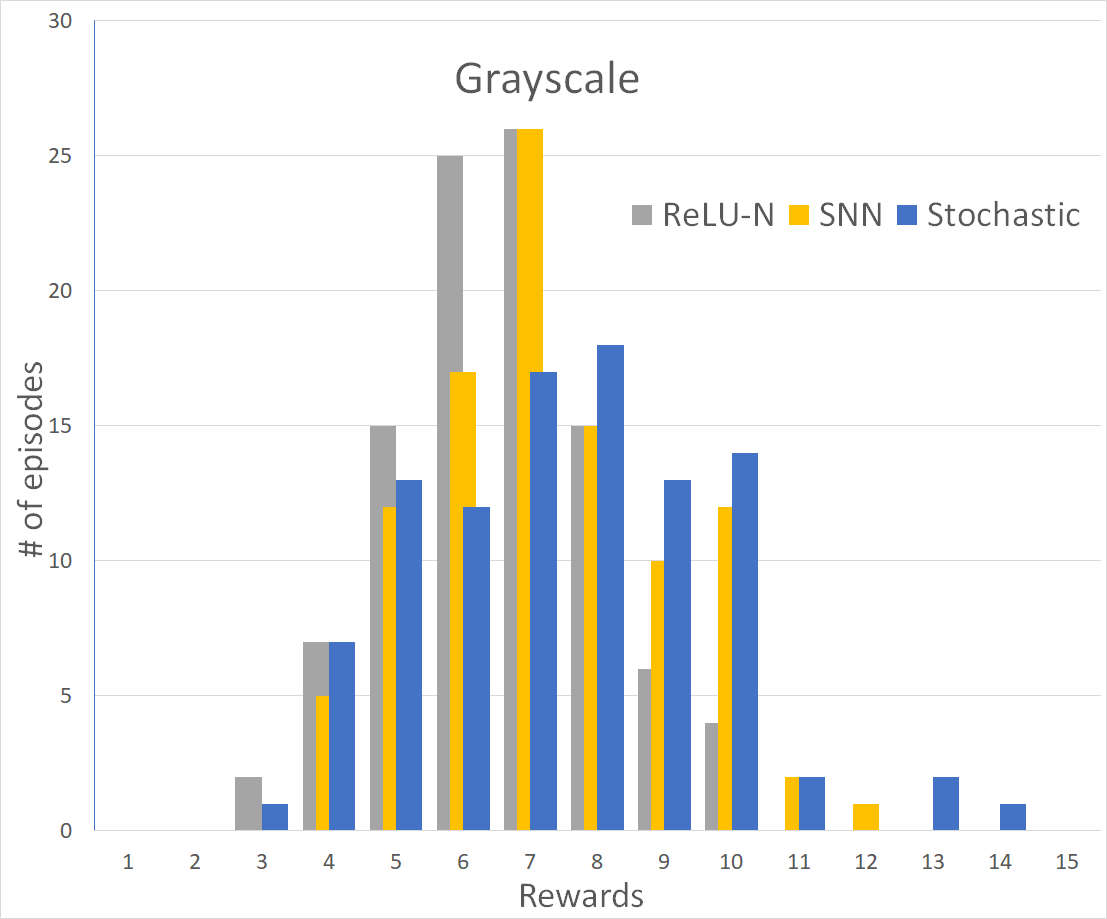}}
\caption{0.05 epsilon greedy grayscale input}
\end{subfigure}
\begin{subfigure}[b]{0.4\linewidth}
\centerline{\includegraphics[width=\linewidth]{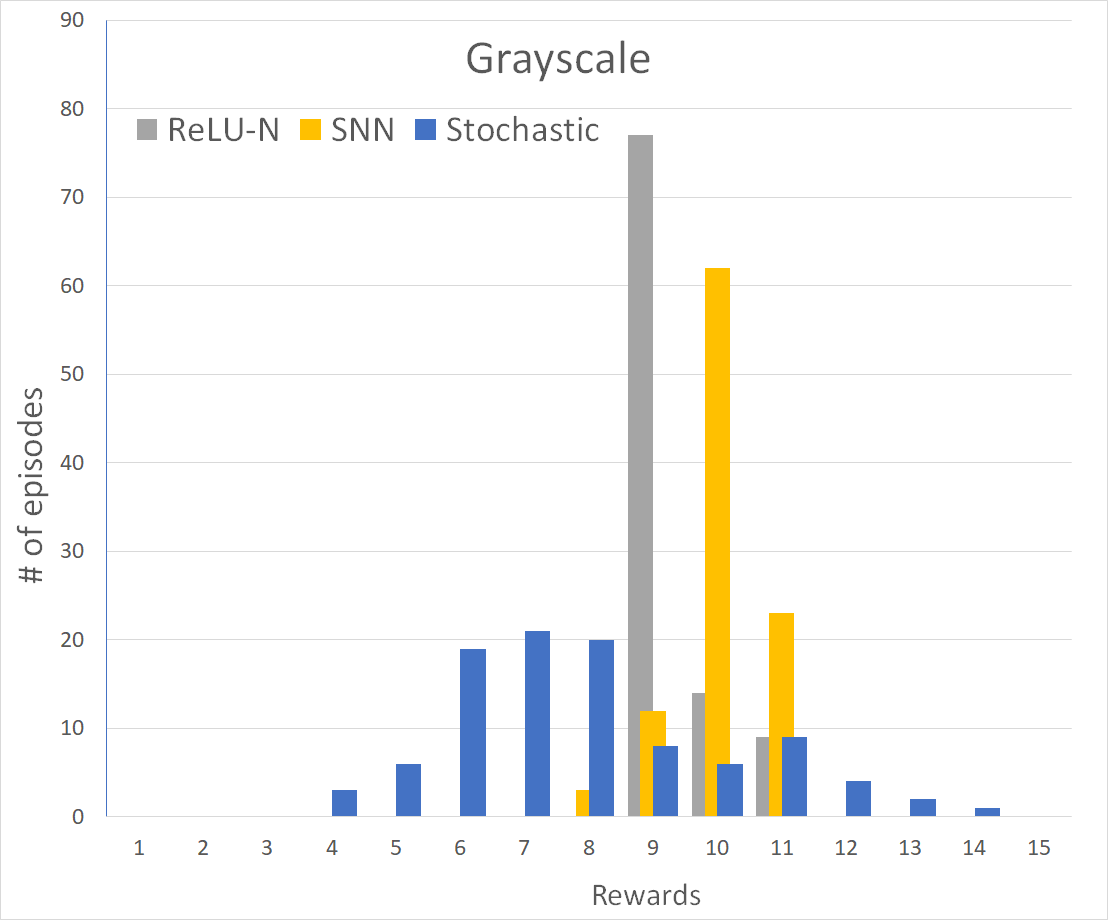}}
\caption{Greedy grayscale input}
\end{subfigure}

\caption{Performance of the networks for Binary and Grayscale inputs; each plot shows the reward distribution over 100 episodes using 0.05 epsilon greedy policy.}
\label{fig:binaryandgrayscaleresults}
\end{center}
\vskip -0.2in
\end{figure*}

We used PSO algorithm to determine the scale of each of the two layers; thus the dimension $D$ of the search space is 2. The swarm size $S$ is set using the formula:
\begin{equation*}
    S = 10 + [2\sqrt{D}],
\end{equation*}
where $[u]$ is the integral part of the real number $u$. For our experiment, the swarm size is $S = 13.$ The fitness of each particle was given by the average reward over 100 episodes.
The stochastic LIF network has a smoother surface of performance over the parameter space than the LIF network. This suggests that the stochastic LIF network is more robust to change in the scaling of its weights. The escape noise of the stochastic LIF neuron can be tuned to improve the performance further however we leave that to future work.

Results of the experiments with shallow NNs are summarized in Table \ref{table:final}. The displayed performance values are obtained by running 100 episodes using two different input encoding (Binary and Grayscale), and applying two policies (greedy and 0.05 epsilon-greedy). We tested the ReLU NN, SNN using LIF neurons, and SNNs using stochastic LIF neurons. 

Table \ref{table:final} summarizes the performance of the ReLU NN against the performance of SNN and the stochastic SNN for binary and grayscale inputs. 
Data in Table \ref{table:final} with \textit{Binary input} demonstrate that SNNs are capable of representing policies developed through RL, and they can outperform the ReLU NNs they originate from. We can see that the stochastic SNN performs better on average than the ReLU NN it has been converted from. The optimal parameters for the binary input spiking neural networks were found using grid search.

Table \ref{table:final} shows that the performance of networks with \textit{Grayscale input} is higher then the \textit{Binary input} for both networks (SNN and ReLU NN). Note that the shown reward values significantly exceed the values by random choice, which is 1.27 $\pm$ 1.45 in these experiments. Note that the ANN has no probabilistic components, and by starting the games with the same initialization over 100 episodes, it reproduces the same outcome every time (zero standard deviation).We also see that the standard deviation of the rewards gained by the SNN is lower and the behavior is less random than for the binary input.

Figure \ref{fig:binaryandgrayscaleresults} provides further details on the performance of the various classification methods, using the histograms of the the reward distributions. The distributions in Fig.  were determined using 0.05 epsilon greedy policy with binary and grayscale inputs. 

\subsection{Robustness of the SNN performance}

Deep Q-networks are vulnerable to white-box and black-box adversarial attacks \citep{Huang2017AdversarialAO}. Witty et al.\citep{witty2018measuring} showed that the policies learned by the DQN algorithm generalize poorly to the states that the agent has not seen during training. To evaluate the robustness of the SNN, we test the performance of the shallow Relu-N and SNN networks with grayscale input when a 3-pixel thick horizontal bar spanning the entire width of the input is occluded. The thickness of the occlusion bar corresponds to the thickness of the paddle on the screen after preprocessing. We tested the performance for every position of the bar, by moving it from the lowest position at the bottom of the screen, step-by-step until it reaches the top. The position of the occlusion bar does not change during each episode. This is a challenging task, since the bar may completely or partially occlude the the ball or the paddle. 

\begin{figure}[ht]
    \begin{center}
        \begin{subfigure}[b]{\columnwidth}
        \centerline{\includegraphics[width=8cm]{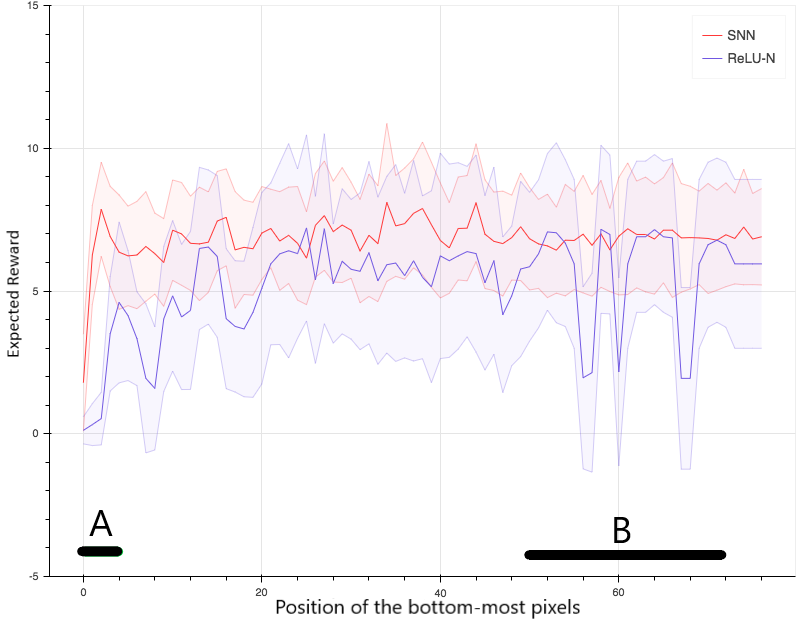}}
        \caption{Pixel-wise robustness ReLU-N vs SNN}
        \end{subfigure}
        \caption{Performance of ReLU-N and SNN for the robustness test. The x-axis represents the position of the bottom most occluded pixels of the 3-pixel thick horizontal occlusion bar. The y-axis represents the average reward. The standard distribution for the reward distribution is shown using the shaded region. The two critical areas are marked by the black bars \textbf{A} and \textbf{B} at the bottom of the plot. \textbf{A} shows the area near the paddle, while \textbf{B} marks the region of the screen occupied by the brick wall.}
        \label{fig:robustness}
    \end{center}
\end{figure}

Figure \ref{fig:robustness} shows the performance of the ReLU NN and SNN for the robustness task. The x-axis represents the vertical position of the lowest occluded pixels. As we move from left to right on the plot, the occlusion bar moves from bottom to the top of the screen; this represent in total 77 experiments for the 77 positions of the occlusion bar. Each experiment was run for 100 episodes using 0.05 epsilon greedy policy.

The SNN is more robust to occlusions than ReLU NN, as it is seen in Fig.  \ref{fig:robustness}, as the reward of the SNN (red) is typically higher that the reward of ReLU NN (blue). 
Moreover, the ReLU NN is very sensitive to occlusions and perturbations at a few places in the input; namely at the bottom near the paddle (shows as region $A$), and at the medium positions where the brick wall is located (region $B$). When these areas are occluded, the ReLU-N performs poorly. 
Occlusions in these areas result in drastic decrease in the performance of the ReLU NN. 
Occluding some of the positions in area $B$, results in a sharp drop in performance for ReLU NN. This can be explained by the nature of the gradient descent updates. Since the score changes when the ball hits the bricks and the backpropagation loss calculated using the TD-error is highest when the score changes, the filters of the network learn to discriminate these areas. Thus, when these areas are occluded, the performance drops. 
Occlusions near areas $A$ and $B$ have much less negative impact on the performance of the SNN. Once the paddle is visible, we see that the SNN has no significant loss in performance. Over the other sensitive occlusion area $B$ near the wall, where ReLU NN has significant drop in performance, SNN performance is sustained without deterioration.
For detailed list of results for positions of the occlusion see \textit{Supplementary Materials}, Table 3. 
We are intensively working on the interpretation of this robustness result and its generalization to a range of task domains.

\section{Performance of SNN obtained by weight transfer from Deep Q-Network}

\begin{figure}[ht]
\vskip 0in
\begin{center}
\centerline{\includegraphics[width=\columnwidth]{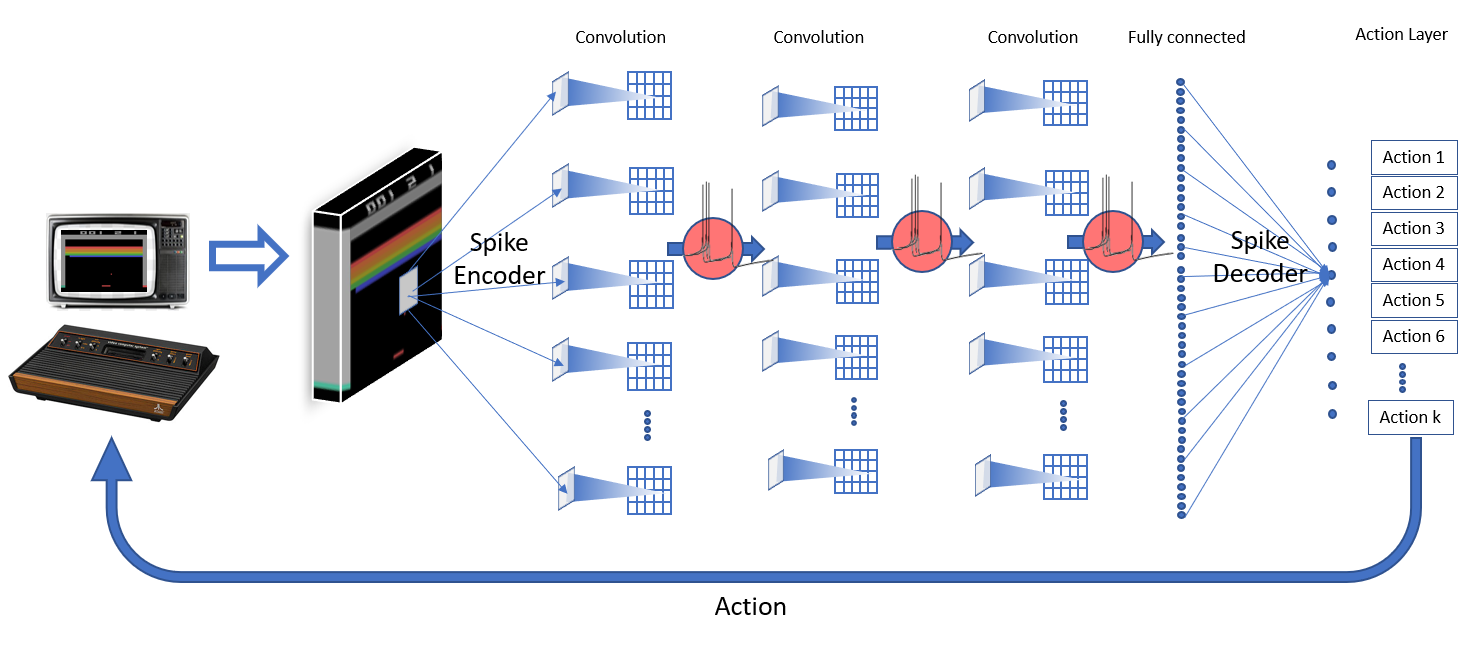}}
\caption{Network architecture following \citep{mnih2015humanlevel}, after converting ReLU
 nonlinearity to spiking network. }
\label{fig:Full_Size_SCNN}
\end{center}
\vskip -0.2in
\end{figure}

To test our approach for state-of-the-art, large-scale networks, we trained the Deep Q-Network \citep{mnih2015humanlevel} and converted the weights to SNN. We used the OpenAI baseline implementation of DQN to train the network \citep{baselines}.
Figure \ref{fig:Full_Size_SCNN} shows the SNN converted from the Deep Q-Network in Figure \ref{fig:Full_Size_CNN}.
Since converting the DQN to SNN requires a search for a large number of parameters, we used the established parameter normalization method \citep{10.3389/fnins.2017.00682}. 
This approach shows reasonable performance, although it can be clearly improved using a systematic parameter optimization method, like PSO. In the Deep Q-SNN, we used the subtractive-IF neuron.

\begin{figure}[ht]
\vskip 0in
\begin{center}
\begin{subfigure}[b]{\columnwidth}
\centerline{\includegraphics[width=8cm]{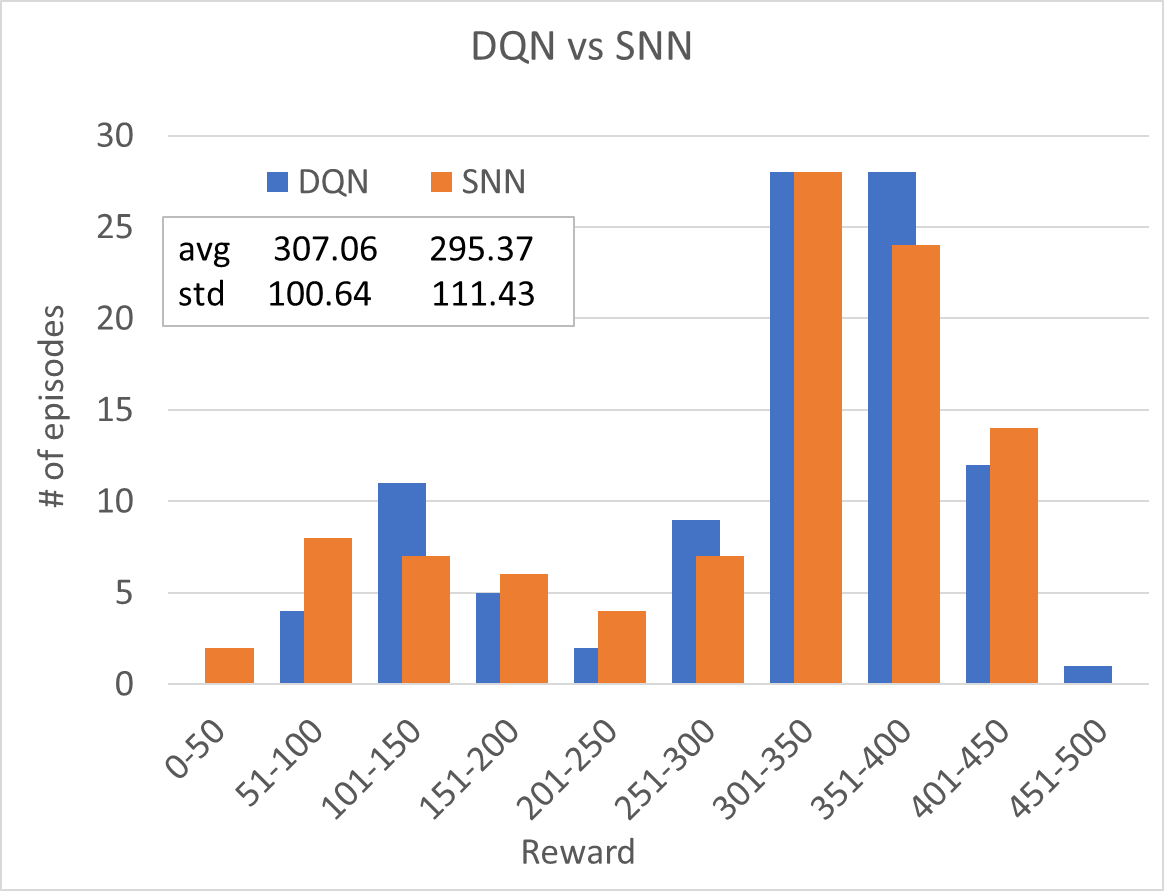}}
\caption{DQN vs SNN}
\end{subfigure}
\caption{Performance of Deep Q-network vs. Deep Spiking Network. Each plot shows the reward distribution over 100 episodes using 0.05 epsilon greedy policy.}
\label{fig:DeepCNNvsSNN_Performance}
\end{center}
\vskip -0.2in
\end{figure}

Figure \ref{fig:DeepCNNvsSNN_Performance} displays the distribution of the rewards in the DQN and Deep SNN. These results show that the Deep Q-Network can be converted to Spiking Q-network without significant loss in performance. At the present stage of our studies, we did not conduct robustness test for the full-scale trained networks. We leave a systematic robustness study and comparison as the objective of future research.

\section{Conclusions}
In this paper, we demonstrate that shallow and deep ReLU NNs trained on the game breakout can be converted to SNNs without degradation of performance. Moreover, SNN seems to display more robustness to occlusion attack. We hypothesize that robustness may be due to the binary nature of spiking neurons, ignoring small perturbations in the data unlike high-precision traditional neural networks. Moreover, the properly optimized conversion method from ReLU to spiking nonlinearity also contributes to the robustness of the results. Moreover, in some cases, SNNs perform better than ReLU NN on previously unseen states. 

These results, combined with additional benefits of SNNs, such as energy efficiency on neuromorphic hardware, show that SNNs may be useful to supplement the power of reinforcement learning in DQN tasks, when resources are limited and the input data are noisy and potentially misleading.

\section*{Acknowledgements}
The work of D.P., H.H., D.J.S., and R.K. has been supported in part by Defense Advanced Research Project Agency Grant, DARPA/MTO HR0011-16-l-0006. Partial support has been provided to R.K. by National Science Foundation Grant NSF-CRCNS-DMS- 13-11165. H.T.S. contribution to this work took place prior to assuming her position at DARPA. The information contained in this work does not necessarily reflect the position or the policy of the Government. 

\section{References}

\bibliographystyle{IEEEtran}
\bibliography{NNDevdharRev2.bib}

\end{document}